\documentclass[11pt]{article}
\usepackage{coling2020}
\usepackage{times}
\usepackage{url}
\usepackage{tablefootnote}
\usepackage{latexsym}
\usepackage{graphicx}

\colingfinalcopy 

\title{KUISAIL at SemEval-2020 Task 12: BERT-CNN for Offensive Speech Identification in Social Media}
 
\author{
  Ali Safaya, \ Moutasem Abdullatif, \ Deniz Yuret \\
  KUIS AI Lab \\
  Ko\c{c} University, Istanbul/Turkey \\ 
  {\tt \{asafaya19,mabdullatif18,dyuret\}@ku.edu.tr}
  }

\date{}

\begin{document}
\maketitle
\begin{abstract}

In this paper, we describe our approach to utilize pre-trained BERT models with Convolutional Neural Networks for sub-task A of the Multilingual Offensive Language Identification shared task (OffensEval 2020), which is a part of the SemEval 2020. We show that combining CNN with BERT is better than using BERT on its own, and we emphasize the importance of utilizing pre-trained language models for downstream tasks. Our system, ranked 4\textsuperscript{th} with macro averaged F1-Score of 0.897 in Arabic, 4\textsuperscript{th} with score of 0.843 in Greek, and 3\textsuperscript{rd} with score of 0.814 in Turkish. Additionally, we present ArabicBERT, a set of pre-trained transformer language models for Arabic that we share with the community.

\end{abstract}

\section{Introduction}

Recently, we have observed an increase in social media usage and a similar increase in hateful and offensive speech. Solutions to this problem vary from manual control to rule-based filtering systems; however, these methods are time-consuming or prone to errors if the full context is not taken into consideration while assessing the sentiment of the text \cite{saif2016}. 

In Subtask-A of the shared task of Multilingual Offensive Language Identification (OffensEval2020), we focus on detecting offensive language on social media platforms, more specifically, on Twitter. The organizers provided data from five different languages, which we worked on three languages of them, namely, Arabic \cite{mubarak2020arabic}, Greek \cite{pitenis2020}, and Turkish \cite{coltekikin2020}. More details about the annotation process have been described in task description paper \cite{zampieri-etal-2020-semeval}.

The approach used combines the knowledge embedded in pre-trained deep bidirectional transformer BERT \cite{devlin-etal-2019-bert} with Convolutional Neural Networks (CNN) for text \cite{kim-2014-convolutional}, which is one of the most utilized approaches for text classification tasks. This combination of models has been shown to yield better results than using BERT or CNN on their own, as was shown in \cite{li-2019-feature-bert}, and shown in this paper. This model, and with minimum text pre-processing,  ranked 4\textsuperscript{th} in Arabic, 4\textsuperscript{th} in Greek, and 3\textsuperscript{rd} in Turkish among more than 40 participants. 

In the following sections of this paper, previous work is mentioned in Section \ref{section1}, next, the data is described in Section \ref{section2}, then the details of the model and the other experiments are described in Section \ref{section3}. Finally, the submissions and the other experiments are detailed in Section \ref{section4}.

\blfootnote{
    \hspace{-0.65cm}  
    This work is licensed under a Creative Commons 
    Attribution 4.0 International License.
    License details:
    \url{http://creativecommons.org/licenses/by/4.0/}.
}

\blfootnote{
    \hspace{-0.65cm} 
    Our source code of the main model and the other experiments can be accessed through: \url{https://github.com/alisafaya/OffensEval2020}
}

\section{Background}\label{section1}
Extensive work has been performed to solve the task of offensive speech identification, which classifies among text classification tasks. Approaches to solve this problem vary from using lexical resources, linguistic features, and meta information \cite{schmidt2017survey}, to machine learning (ML) models \cite{davidson2017automated}, and more recently, deep neural models like CNN and Long-Short Term Memory (LSTM) and their derivatives \cite{zhang2018detecting}.

More recent work, \newcite{zampieri2019predicting} presented Offensive Language Identification Dataset (OLID), which is a new dataset with tweets annotated for offensive content, they experiment with various ML models, like SVM, BiLSTM and CNN.

\section{Data}\label{section2}

The data provided for this task \cite{zampieri-etal-2020-semeval} consists of sets of tweets which were annotated as either \textbf{Offensive} (positive) or \textbf{Non-offensive} (negative). As shown in Table \ref{data-table} below, each set contains a number of positive tweets and negative tweets. In addition, the provided training data had not been split into training and development sets, so the data was split into 90\% and 10\% for training and development sets respectively.

\begin{table}[h]
\begin{center}
\begin{tabular}{|l|lll|lll|lll|} 
\hline
& \multicolumn{3}{c}{\bf Arabic} & \multicolumn{3}{|c|}{\bf Greek} & \multicolumn{3}{c|}{\bf Turkish} \\
& Train & Dev & Test & Train & Dev & Test & Train & Dev & Test \\
\hline
\bf Negative & 5,785 & 626 & 1,607 & 5,642 & 616 & 1,120 & 23,084 & 2,543 & 2,740 \\ 
\bf Positive & 1,415 & 174 & 393 & 2,228 & 258 & 424 & 4,885 & 632 & 788 \\ \hline
\bf Total & 7,200 & 800 & 2,000 & 7,869 & 874 & 1,545 & 28,581 & 3,175 & 3,528 \\ \hline

\end{tabular}
\end{center}
\caption{\label{data-table} Tweets distribution over data sets }
\end{table}

\subsection{Data Pre-processing}

Since processed texts were obtained from Twitter, a pre-processing step was needed to maximize the features that can be extracted and to obtain a clean text; Hashtags were converted into raw text by splitting the texts into words, for example: \url{#SomeHashtagText} becomes \url{Some} \url{Hashtag}  \url{Text}. As an additional step only for Greek texts, all letters to converted to lowercase letters and all Greek diacritics were removed. The text was subsequently tokenized using the corresponding BERT pre-trained Wordpiece tokenizer for each language and model.

\section{Model Description}\label{section3}

The proposed model maximizes the utilization of knowledge embedded in pre-trained BERT language models by feeding the outputted contextualized embeddings of its last four hidden layers into a several filters and convolution layers of the CNN. Finally, the output of the CNN was passed to a dense layer and the predictions were obtained.

\subsection{Convolutional Neural Networks}

CNN for textual tasks by \newcite{kim-2014-convolutional} showed superiority in text classification tasks. CNNs can be used with learned vector representations of the text (embeddings). These embeddings may either be initialized randomly and trained along with the model, or can be pre-trained vectors.

\subsection{BERT}

Bidirectional Encoder Representations from Transformers (BERT) \cite{devlin-etal-2019-bert} is state-of-the-art language model, which can be fine-tuned, or used directly as a feature extractor for various textual tasks. In our experiments, three pre-trained language-specific BERT models were used along with Multilingual-BERT (mBERT)\footnote{Multilingual: \url{https://github.com/google-research/bert}} model. Those models are GreekBERT\footnote{GreekBERT: \url{https://github.com/nlpaueb/greek-bert}} model for Greek, BERTurk \cite{stefan-bert} for Turkish, and ArabicBERT for Arabic.

\subsection{ArabicBERT}

Since there was no pre-trained BERT model for Arabic at the time of our work, four Arabic BERT language models were trained from scratch and made publicly available for use. 

\textbf{ArabicBERT}\footnote{ArabicBERT: \url{https://github.com/alisafaya/arabic-bert}} is a set of BERT language models that consists of four models of different sizes trained using masked language modeling with whole word masking \cite{devlin-etal-2019-bert}. Models of sizes \textbf{Large}, \textbf{Base}, \textbf{Medium}, and \textbf{Mini} \cite{turc2019well} were trained on the same data for 4M steps.

Using a corpus that consists of the unshuffled version of OSCAR data \cite{ortiz-suarez-etal-2020-monolingual} and a recent data dump from Wikipedia, which sums up to 8.2B words, a vocabulary set of 32,000 Wordpieces was constructed. The final version of corpus contains some non-Arabic words inlines, which were not removed from sentences since that would affect some tasks like Named Entity Recognition. Although non-Arabic characters were lowered as a pre-processing step, since Arabic characters do not have upper or lower case, there is no cased and uncased version of the model. Subsequently, the corpus and the vocabulary set are not restricted to Modern Standard Arabic, they contain some dialectical (spoken) Arabic too, which boosted models performance in terms of data from social media platforms.

\begin{figure}
    \centering
    \includegraphics[scale=0.4]{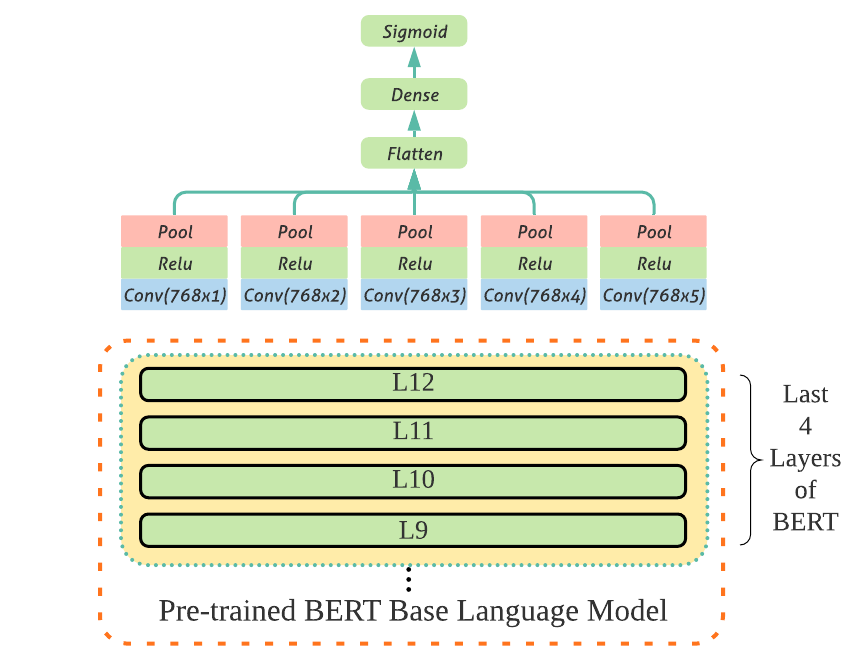}
    \caption{BERT-CNN model structure}
\end{figure}

\subsection{BERT-CNN Model Structure}


As mentioned above, the main model consists of two main parts. The first part being BERT Base model, in which the text is passed through 12 layers of self-attention to obtain contextualized vector representations. The other part being CNN, which was used as a classifier.

\newcite{devlin-etal-2019-bert} showed by comparing different combinations of layers of BERT, that the output of the last four hidden layers combined, encodes more information than the output of the top layer. 

After setting the maximum sequence length of each text sample (tweet) to 64 tokens, the text was input to BERT, then the output of the last four hidden layers of base sized pre-trained BERT, was concatenated to get vector representations of size 768x4x64 as shown in Figure 1. Next, these embeddings were passed in parallel into 160 convolutional filters of five different sizes (768x1, 768x2, 768x3, 768x4, 768x5), 32 filters for each size. Each kernel takes the output of the last four hidden layers of BERT as 4 different channels and applies convolution operation on it. After that, the output is passed through ReLU Activation function and a Global Max-Pooling operation. Finally, the output of the pooling operation is concatenated and flattened to be later on passed through a dense layer and a Sigmoid function to get the final binary label.

This model was trained for 10 epochs with learning rate of 2e-5, and the model with the best macro averaged F1-Score on the development set was saved.

\section{Experiments and Results}\label{section4}

\begin{table}[h]
\begin{center}
\begin{tabular}{|c|c|c|c|c|} 
\hline
\bf Model & \bf Arabic & \bf Greek & \bf Turkish & \bf Average \\
\hline
\bf SVM with TF-IDF & 0.772 & 0.823	& 0.685 & 0.760 \\ 
\bf Multilingual BERT & 0.808 & 0.807 & 0.774 & 0.796 \\
\bf Bi-LSTM & 0.822 & 0.826 & 0.755 & 0.801 \\
\bf CNN-Text & 0.840 & 0.825 & 0.751 & 0.805 \\
\bf BERT\tablefootnote{\label{bert-note}Language specific pre-trained BERT models were used for this experiment} & 0.884 & 0.822 & \bf0.816 & 0.841 \\
\bf BERT-CNN (Ours)\footnotemark[4] & \bf0.897 & \bf0.843 & 0.814 & \bf0.851 \\
\hline

\end{tabular}
\end{center}
\caption{\label{results-table} Macro averaged F1-Scores of our submissions and the other experiments on test data}
\end{table}

Macro-averaged F1-Score metric was used for evaluation in this shared task. The results of our submissions were shown in comparison with other experiments in Table \ref{results-table}.

These experiments began by building a baseline model using a classic ML approach for text classification. Additionally, the main model was compared with more recent approaches. All the models use the same train/dev/test splits.

\subsection*{SVM with TF-IDF\footnote{This model was built using Scikit-learn: \url{scikit-learn.org}}}

The baseline model used Term Frequency-Inverse Document Frequency (TF-IDF) \cite{Salton1988} with Support Vector Machine (SVM) \cite{Boser92atraining}. Count Vectorizer with feature set size of 3000 was used to achieve the results demonstrated in Table \ref{results-table}.

\subsection*{CNN-Text\footnote{\label{pytorch} This model was built using PyTorch: \url{pytorch.org}}} Using CNNs with the same structure as the main model, but without pre-trained BERT as an embedder. CNN-Text model uses randomly initialized embeddings of size 300, which were trained along with the model. The difference between the results obtained using pre-trained BERT and randomly initialized embeddings was significant as shown in the Table \ref{results-table} above.

\subsection*{BiLSTM\footnotemark[6]}

While CNNs could be used to capture local features of the text, LSTM which have shown remarkable performance in text classification tasks, capture the temporal information. In our experiments, two layers of Bidirectional LSTM (BiLSTM) with a hidden size of 128, and randomly initialized embeddings of size 300, were used to achieve the results shown in Table \ref{results-table}, However, this was still outperformed by CNN-Text on average.

\subsection*{BERT\footnotemark[6]\textsuperscript{,}\footnote{Transformers library was used for BERT \cite{Wolf2019HuggingFacesTS}}}

By looking at the average results of BERT model on its own, we can see the improvement that was achieved by combining BERT with CNN. Additionally we can clearly observe the advantage of using \textbf{Language-specific} pre-trained models on \textbf{Multilingual} ones.

\section{Conclusion}
In this paper, the structure of BERT-CNN was described and compared with other models on the ability to identify offensive speech text in social media. It was shown that combining BERT with CNN yields better than using BERT on its own. Additionally, the pre-training process of ArabicBERT was explained.
The proposed model with minimum text pre-processing was able to achieve very good results on average and our team was ranked among the highest four participating teams for all languages in the scope of the OffensEval2020.

\section*{Acknowledgements}

The hardware infrastructure of this study is provided by the European Research Council (ERC) Starting Grant 714868. Also, we would like to thank Google for providing free TPUs and Credits for the pre-training process of ArabicBERT and for Huggingface.co for hosting these models on their servers.

\bibliographystyle{coling}
\bibliography{coling2020}

\end{document}